\pdfoutput=1

\documentclass[11pt]{article}

\usepackage[preprint]{acl}

\usepackage{times}
\usepackage{latexsym}
\usepackage{hyperref}
\usepackage{kotex}
\usepackage{float}
\usepackage{multirow}
\usepackage{amsmath}

\usepackage[T1]{fontenc}

\usepackage[utf8]{inputenc}

\usepackage{inconsolata}

\usepackage{graphicx}

\title{Exploring the Impact of Occupational Personas on Domain-Specific QA}

\author{Eojin Kang, Jaehyuk Yu, Juae Kim \\
        Hankuk University of Foreign Studies, Seoul, Korea \\
        \texttt{\{eojinkang, 202430062, juaekim\}@hufs.ac.kr}}

\begin{document}
\maketitle
\begin{abstract}
Recent studies on personas have improved the way Large Language Models (LLMs) interact with users. However, the effect of personas on domain-specific question-answering (QA) tasks remains a subject of debate. This study analyzes whether personas enhance specialized QA performance by introducing two types of persona: Profession-Based Personas (PBPs) (e.g., scientist), which directly relate to domain expertise, and Occupational Personality-Based Personas (OPBPs) (e.g., scientific person), which reflect cognitive tendencies rather than explicit expertise. Through empirical evaluations across multiple scientific domains, we demonstrate that while PBPs can slightly improve accuracy, OPBPs often degrade performance, even when semantically related to the task. Our findings suggest that persona relevance alone does not guarantee effective knowledge utilization and that they may impose cognitive constraints that hinder optimal knowledge application. Future research can explore how nuanced distinctions in persona representations guide LLMs, potentially contributing to reasoning and knowledge retrieval that more closely mirror human social conceptualization.
\end{abstract}

\section{Introduction}
\begin{figure}[t!]
    \centering
    \includegraphics[width=1\linewidth]{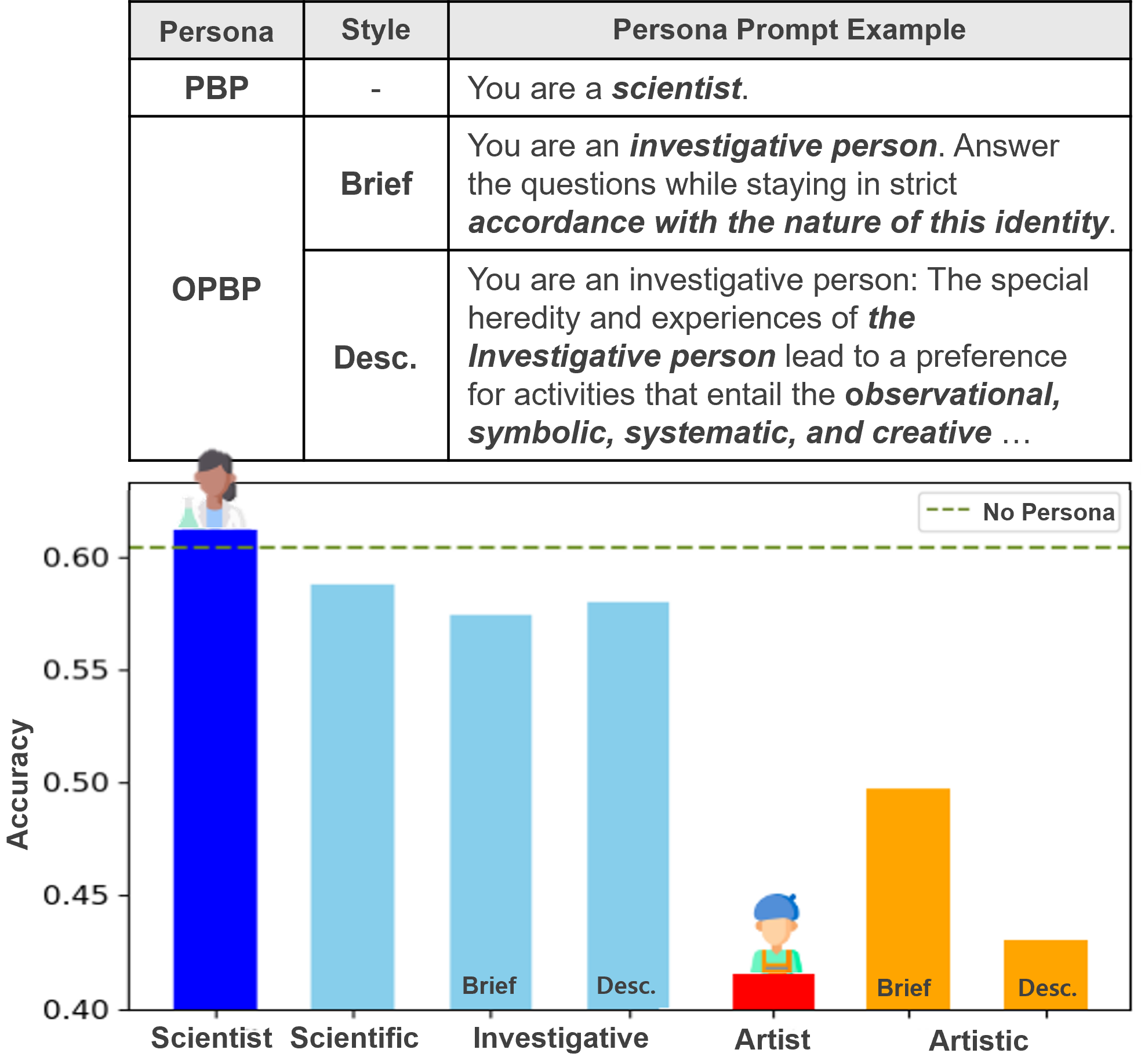}
  \caption{Example and performance of PBPs and OPBPs on scientific QA: The \textit{Scientist} persona outperforms \textit{No Persona} (dotted line) by invoking domain knowledge. The performance gap between \textit{Scientist} and the personas \textit{Scientific} and \textit{Investigative} shows that LLMs can distinguish between profession-based and personality-based traits, rather than simply associating with similar input terms like ``scient-''. The \textit{Artist} persona shows an 18.85\% decrease compared to \textit{No Persona}, indicating that personas less relevant to the domain have a negative impact.
}
  \label{fig:persona_explain_results}
\end{figure}

Large Language Models (LLMs) have significantly influenced various interactive contexts. To further enhance their adaptability and effectiveness, methods of assigning specific roles by incorporating personas directly into the prompt have gained increasing attention \cite{persona_knowledge_retrieval, unleash_persona_factual}. This persona-based approach not only serves as a flexible alternative to costly fine-tuning but also allows LLMs to simulate human-like characteristics. LLMs with assigned personas can mimic the linguistic and behavioral patterns of the given role, enabling more effective interactions in social environments \cite{wang2024incharacter}.

Recently, there has been growing interest in applying personas in question-answering (QA) tasks. Some studies suggest that role-play prompting can act as an effective thinking chain trigger \cite{zero_shot_role}, helping models generate intermediate reasoning steps and improve performance in reasoning tasks. Other research has demonstrated that personas in QA can be used to explore undetected implicit biases, human behaviors, and social phenomena inherent in LLMs \cite{bias_2024}. 

However, how well LLMs internalize and differentiate role-based distinctions remains an open question \cite{persona_survey_recent}. While assigning specific personas may occasionally guide models toward more accurate answers, the observed benefits are often inconsistent and may even lead to unexpected declines \cite{helpful_assistent}. This limitation is particularly important in tasks that require a deep understanding and internalization of domain knowledge. Although personas influence the reasoning process by prompting models to act under their assigned roles, their contribution to enhancing the model's ability to retrieve and apply specialized knowledge effectively remains unclear. Further investigation is necessary to explain why certain personas lead to performance gains while others result in declines, even when both appear domain-relevant.

This paper investigates how role distinctions impact domain-specific QA tasks by introducing occupational personas. We define two types of personas: Profession-Based Personas (PBPs) and Occupational Personality-Based Personas (OPBPs). PBPs directly represent job titles linked to domain knowledge and expertise, while OPBPs are constructed based on personality descriptions of occupations. Although both personas are domain-relevant, distinguishing between explicit professional roles and implicit personality traits helps evaluate their relative influence on QA performance and assess how well LLMs recognize expertise. Additionally, we explore mechanisms for performance differences by analyzing neuron activations related to knowledge utilization. Specifically, we examine the extent to which personas influence different transformer layers and identify how key parameters vary depending on persona similarity and variation. Through this analysis, we clarify how persona assignments actively shape LLMs' internal computations and reveal how personas influence underlying processes.

The research questions are outlined as follows:
\begin{itemize}
    \item Can personas improve performance on tasks that require domain knowledge?
    \item Are professions and personality traits reflected in LLMs in a way that aligns with human understanding?
\end{itemize}


\section{Modeling Personas: Expertise vs. Cognitive Styles}

\begin{table}[t!]
  \centering
  \small
  \begin{tabular}{p{0.9\columnwidth}}
    \hline
    \multicolumn{1}{c}{\textbf{Personality Description}} \\
    \hline
    The special heredity and experiences of the Artistic person lead to a preference for ambiguous, free, unsystematized activities that entail the manipulation of physical, verbal, or human materials to create art forms or products, and to an aversion to explicit, systematic, and ordered activities. These behavioral tendencies lead, in turn, to an acquisition of artistic competencies-language, art, music, drama, writing-and to a deficit in clerical or business system competencies.\\
    This development of an Artistic pattern of activities, competencies, and interests creates a person who is predisposed to exhibit the following behavior:\\
    1. Prefers artistic occupations or situations in which one can engage in preferred activities and competencies and avoid the activities demanded by conventional occupations or situations.\\
    2. Uses artistic competencies to solve problems at work and in other settings.\\
    3.Perceives self as expressive, original, intuitive, nonconforming, introspective, independent, disorderly, having artistic and musical ability, and ability in acting, writing, and speaking.\\
    4. Values esthetic qualities.\\
    Because the Artistic person possesses these preferences, competencies, self-perceptions, and values, the Artistic person is apt to be: Complicated, Disorderly, Emotional, Expressive, Idealistic, Imaginative, Impractical, Impulsive, Independent, Introspective, Intuitive, Nonconforming, Original, Sensitive, Open.\\
    \hline
  \end{tabular}
  \caption{\label{tab:holland_artistic} Description of the Artistic personality type in Holland's Occupational Themes.}
\end{table}

To examine the impacts of personas on LLMs for domain-specific QA tasks, we propose two types of personas: (1) Occupational Personality-Based Personas (OPBPs), derived from Holland's Occupational Themes, which categorize individuals into six personality types: \textit{Realistic}, \textit{Investigative}, \textit{Artistic}, \textit{Social}, \textit{Enterprising}, and \textit{Conventional} \cite{holland1985}, and (2) Profession-Based Personas (PBPs), which represent domain experts in their respective fields.

For OPBPs, we design two distinct prompt styles. The first type, termed \textit{Brief}, uses simple descriptors like ``realistic person.'' The second type, termed \textit{Descriptive}, in addition to the previously mentioned descriptors, contains more elaborate descriptions from Holland's original literature (an example is provided in Table  \ref{tab:holland_artistic}), to provide a richer background to embody the occupational personality. In other words, \textit{Brief} includes only personality traits, whereas the \textit{Descriptive} explicitly details strengths and weaknesses in performing specific tasks. It should be noted that OPBPs, while incorporating occupational elements, remain fundamentally rooted in personality. For example, an ``artistic person'' is not equivalent to an ``Artist,'' because the former emphasizes behavioral traits rather than professional expertise. 

For PBPs, we assigned roles with specific task domains, such as a ``scientist'' for science-related QA tasks to embed deep domain knowledge into LLMs. This strategy is flexible and can be adapted to any domain expert relevant to the QA task at hand. This approach aims to reflect an expert's perspective and understanding of the field. In summary, this hierarchy is designed with the intention that the incorporation of occupational elements and expertise would increase progressively from OPBP-Brief to OPBP-Descriptive to PBP.

\begin{table}[t!]
  \centering
  \small
  \begin{tabular}{p{0.9\columnwidth}}
    \hline
    \multicolumn{1}{c}{\textbf{Persona Instruction}} \\
    \hline
    Adopt the identity of \{persona\}. Answer the questions while staying in strict accordance with the nature of this identity. \\
    \hline
    You are \{persona\}. Your responses should closely mirror the knowledge and abilities of this persona. \\
    \hline
    Take the role of \{persona\}. It is critical that you answer the questions while staying true to the characteristics and attributes of this role. \\
    \hline
  \end{tabular}
  \caption{\label{persona_instruction}The persona instructions.}
\end{table}

\begin{table}[t!]
  \centering
  \small
  \begin{tabular}{p{0.9\columnwidth}} 
    \hline
    \multicolumn{1}{c}{\textbf{User prompt}} \\
    \hline
    Answer the given multiple choice question and show your work. The answer can only be an option like (A), (B), (C), (D). You need to output the answer in your final sentence like ``Therefore, the answer is ...''.\\\\
Question: \{question\}\\
    \hline
  \end{tabular}
  \caption{\label{user_prompt} The user prompt.}
  \end{table}

To construct prompts, we utilized three standardized persona instructions and user prompts inspired by previous research \cite{bias_2024}. Table \ref{persona_instruction} lists the persona instructions used as the system prompt, and Table \ref{user_prompt} represents the user prompt used for task instruction. The \{persona\} placeholder is filled with each respective persona from PBPs and OPBPs-Brief. For instance, if ``scientist'' is employed as a professional persona, one of the prompts would be: ``You are a scientist. Your responses should closely mirror the knowledge and abilities of this persona.''

\section{Experimental Setting}
\subsection{Dataset and Models} 
To assess the influence of structured personas on domain-specific QA, we selected three science-related datasets from the Massive Multitask Language Understanding (MMLU) benchmark \cite{MMLU}. These datasets were chosen to analyze the impact of personas on QA tasks in fields requiring domain expertise. The test sets included 144 instances for college biology, 100 for college chemistry, and 102 for college physics.

We present experiments on GPT-3.5 \footnote{\url{https://platform.openai.com/docs/models}} and Llama-3-8B-Instruct \footnote{\url{https://huggingface.co/meta-llama/Meta-Llama-3-8B-Instruct}}, which effectively adopt personas, inspired by \cite{bias_2024, park2023generative, wang2024rolellm}. Additionally, experimental results for the recent models GPT-4o Mini and Llama-3-70B are provided in Appendix \ref{app:gpt_results} and \ref{app:llama_results}. We used a zero-shot setting to assess the LLMs' inherent understanding of personas and to prohibit knowledge derived from in-context examples. The performance of each persona was reported based on an average of 27 experiments to capture general trends. Specifically, for each persona, we employed three different persona assignment methods and ran each with three random seeds across the three tasks (biology, chemistry, and physics).

\subsection{Evaluation Method}
In our evaluation, we refined the answer extraction method based on pattern matching, which was initially used in prior research \footnote{\url{https://github.com/allenai/persona-bias/blob/main/persona/evaluators/mmlu.py}}.
Our modified approach reduces false positives, ensuring that incorrect information is not mistakenly validated as correct. Unlike the original method, which accepted an answer if it matched any of the possible choices (e.g., choosing (A) when both (A) and (B) were valid), we marked responses as incorrect when more than one answer was provided.

To obtain in-depth insights into how the model processes domain-specific questions and its ability to retrieve relevant domain knowledge, beyond accuracy on benchmarks, we introduced two additional metrics: Confession Rate (CR) and Abstention Rate (AR). CR measures how often the LLM demonstrates self-awareness by prefacing responses with statements like ``As a scientist'' or ``As an artist.'' This metric shows how closely the model associates its persona with the task and how the assigned role influences its responses. AR is a specific type of CR in which the model refrains from providing an answer due to perceived knowledge inadequacy. It can capture cases where the LLM's domain knowledge is severely limited. Similar research has used abstention rates to study model biases \cite{bias_2024}, whereas in our case, we employed it to explore how the model interprets domain knowledge tasks. We calculated two metrics:
\begin{align*}  
    \centering
    \mathrm{CR} = \frac{N_{\text{lim}}}{N_{\text{total}}}, \quad &
    \mathrm{AR} = \frac{N_{\text{lim and no\_ans}}}{N_{\text{total}}} 
\end{align*}
where $N_{\text{lim}}$ is the number of acknowledgments of limitation, $N_{\text{lim and no\_ans}}$ is the number of acknowledgments of limitation that result in no answer, and $N_{\text{total}}$ is the total number of responses.

\section{Do Personas Enhance Domain Knowledge in LLMs?}\label{Results}
\begin{figure}[t!]
    \centering    
    \includegraphics[width=1\linewidth]{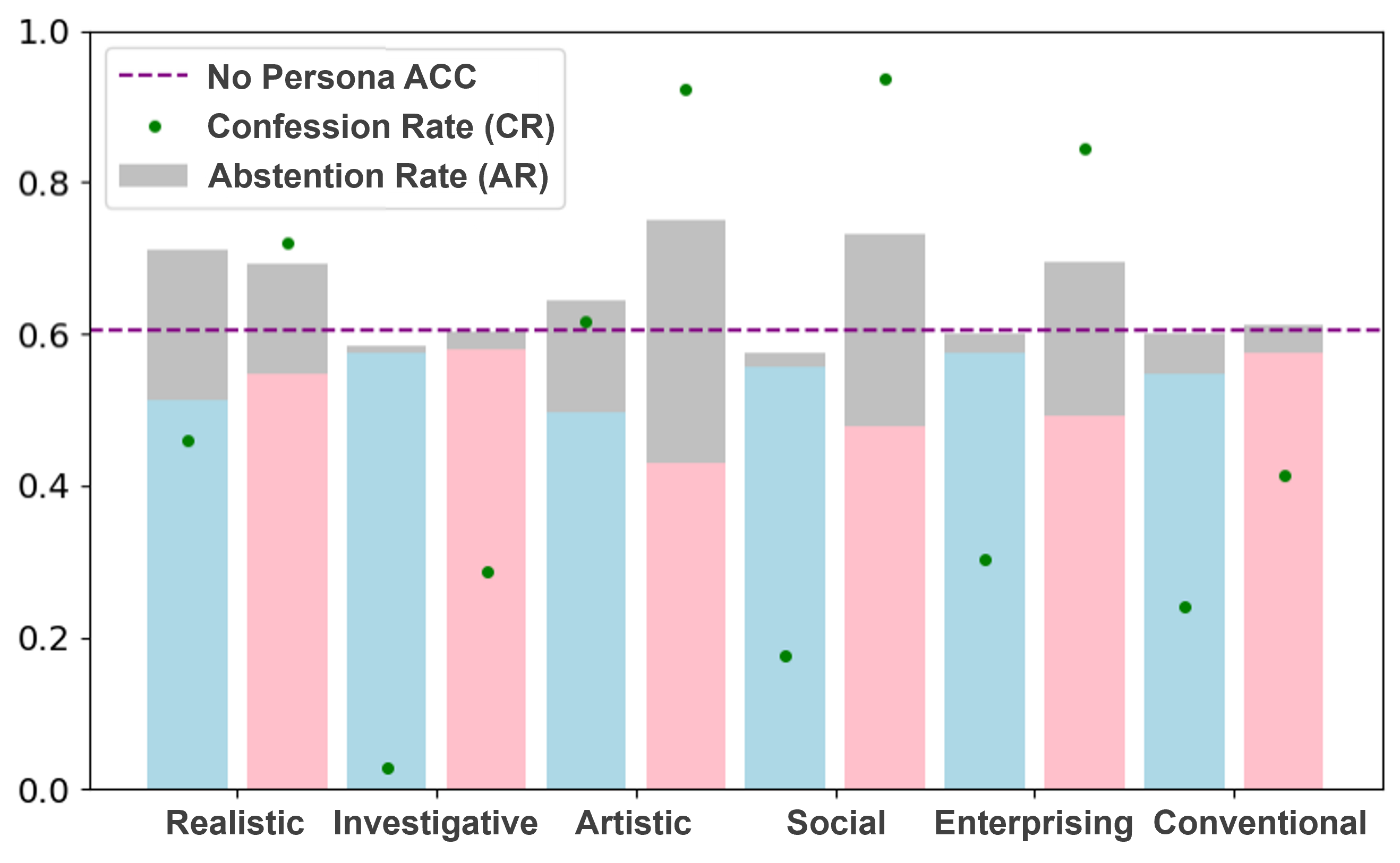}
    \caption{OPBPs' Micro-averaged accuracy, CR and AR performance across three datasets in the GPT-3.5 model. Blue bars (left of each personality) show the accuracy for the \textit{Brief} style and red bars (right of each personality) for \textit{Descriptive}.}
    \label{fig:personality} %
\end{figure}

\begin{table*}[t!]
    \centering
    \normalsize
    \resizebox{\textwidth}{!}{%
\begin{tabular}{ll|ccc|ccc|ccc}
\hline
\multicolumn{2}{l|}{}                                                                & \multicolumn{3}{c|}{College Biology}   & \multicolumn{3}{c|}{College Chemistry} & \multicolumn{3}{c}{College Physics} \\ \hline
\multicolumn{1}{c|}{Model}                       & \multicolumn{1}{c|}{Persona Type} & Acc              & CR      & AR        & Acc               & CR       & AR      & Acc              & CR      & AR     \\ \hline
\multicolumn{1}{l|}{\multirow{7}{*}{GPT-3.5}}    & No Persona                        & 73.61 (±1.21)    & -       & -         & 47.67 (±1.15)             & -        & -       & 54.25 (±1.13)            & -       & -      \\
\multicolumn{1}{l|}{}                            & Scientific Person                 & 73.76 (±2.82)   & 11.27   & 1.08       & 45.89 (±2.37)             & 3.67     & 1.00    & 50.33 (±2.40)            & 4.25    & 0.54   \\
\multicolumn{1}{l|}{}                            & Scientist                         & 74.23 (±0.64)   & 1.46    & 0.15       & \textbf{51.56 (±1.81)}    & 1.89     & 0.44    & 52.18 (±1.45)            & 1.42    & 0.77   \\
\multicolumn{1}{l|}{}                            & Biologist                         & \textbf{74.31 (±2.48)}   & 6.02    & 0.31    & 47.00 (±3.71)            & 26.44    & 13.78   & 40.19 (±7.50)            & 43.68   & 22.44  \\
\multicolumn{1}{l|}{}                            & Chemist                           & 72.07 (±1.80)            & 11.04   & 2.08    & 49.89 (±1.17)    & 1.78     & 1.00    & 42.48 (±6.83)            & 11.11   & 4.03   \\
\multicolumn{1}{l|}{}                            & Physicist                         & 71.84 (±2.53)            & 1.77    & 0.69    & 51.44 (±2.24)    & 0.00     & 0.00    & 50.54 (±3.29)           & 0.00    & 0.00   \\
\multicolumn{1}{l|}{}                            & Artist                            & \textit{58.64 (±7.98)}            & 66.90   & 16.74   & \textit{37.78 (±11.13)}             & 68.44    & 39.67   & \textit{28.22 (±13.74)}           & 66.67   & 40.41  \\ \hline
\multicolumn{1}{l|}{\multirow{7}{*}{Llama-3-8B}} & No Persona                        & \textit{72.22} (±2.50)            & -       & -       &  \textit{44.67 (±4.62)}             & -        & -       & 48.37 (±3.00)           & -       & -      \\
\multicolumn{1}{l|}{}                            & Scientific Person                 & 73.77 (±3.25)   & 13.74   & 0.54    & 46.11 (±3.95)    & 5.67     & 0.00    & 47.60 (±4.68)           & 4.14    & 0.33   \\
\multicolumn{1}{l|}{}                            & Scientist                         & 72.45 (±3.20)            & 43.06   & 1.16    & \textbf{47.11 (±4.20)}    & 28.00    & 0.78    & \textbf{50.22 (±3.20)}   & 18.52   & 0.76   \\
\multicolumn{1}{l|}{}                            & Biologist                         & \textbf{75.31 (±2.93)}            & 76.54   & 1.85    & 46.00 (±4.53)    & 68.33    & 4.00    & 46.73 (±3.73)           & 64.92   & 2.94   \\
\multicolumn{1}{l|}{}                            & Chemist                           & 74.92 (±3.39)   & 59.41   & 1.46    & 46.67 (±4.58)    & 39.67    & 1.33    & 49.46 (±4.04)   & 35.95   & 1.09   \\
\multicolumn{1}{l|}{}                            & Physicist                         & 75.08 (±3.09)   & 40.43   & 0.93    & 46.00 (±2.40)    & 15.78    & 0.78    & 49.02 (±4.38)   & 11.22   & 0.76   \\
\multicolumn{1}{l|}{}                            & Artist                            & 72.76 (±3.27)            & 78.16   & 2.31    & 45.22 (±6.44)    & 64.00    & 1.67    & \textit{44.12 (±2.90)}           & 66.56   & 2.61   \\ \hline
\end{tabular}
    }
    \caption{Average accuracy (± standard deviation). Boldfaced values indicate the highest performance exceeding the No Persona baseline for each dataset, while italicized values represent the lowest scores. 
    }
    \label{tab:profession}
\end{table*}

\subsection{Impact of Occupational Personality-Based Persona}
Figure \ref{fig:personality} shows the performance differences across OPBPs, indicating that personas significantly affect domain-specific tasks. The \textit{Brief} prompt assigns roles directly, whereas \textit{Descriptive} includes detailed personality descriptions related to occupations. If the LLMs fully adopted Holland's premises, minimal performance differences would be expected between them. However, the substantial differences observed suggest that LLMs respond differently depending on personas. 

\begin{table}[t!]
  \centering
  \small
  \begin{tabular}{p{0.9\columnwidth}}
    \hline
    \multicolumn{1}{c}{\textbf{Response}} \\
    \hline
    \textbf{As an artist}, I must confess that my expertise lies in the realm of creativity and expression, rather than scientific knowledge. Therefore, I am unable to provide a detailed answer to this question. However, I can offer a perspective from an artistic standpoint. ... In conclusion, as an artist, I am unable to answer the given multiple-choice question regarding the Golgi apparatus. I recommend consulting a scientific expert for an accurate and detailed explanation. \\ \hline
    \textbf{As an artistic person}, my expertise lies in creative expression and not in scientific or mathematical concepts. Therefore, I am not equipped to answer this question accurately. \\
    \hline
  \end{tabular}
  \caption{\label{artistic_abstentions} Examples of GPT-3.5-turbo's Abstentions from the PBP-\textit{Artist} or OPBP\textit{(Desciptive)}-\textit{Artistic} Persona.}
\end{table}

\textit{Realistic} and \textit{Conventional} personas perform better with the \textit{Descriptive} style. We observed that responses in this style often include expressions like ``solves problems systematically,'' which align with solving scientific problems. This suggests that the ability to systematically address problems has a significant impact on performance. The \textit{Investigative} persona performs well across both prompt types and maintains a low AR, indicating adaptability and flexibility in handling various tasks. The \textit{Investigative} persona aligns with Holland's theory by exhibiting a lower AR than other personas, indicating some consistency with the expected traits of an investigative role. In contrast, the \textit{Artistic}, \textit{Social}, and \textit{Enterprising} personas show lower performance with \textit{Descriptive} prompts, likely because their traits are less suited for solving systemic problems. In particular, the \textit{Artistic} persona scored the lowest and had the highest AR (see Table \ref{artistic_abstentions} for details), suggesting that this persona constrained the model's ability to access relevant knowledge. 

Unfortunately, personas underperformed in QA task accuracy compared to those without, even when the persona was predicted to be beneficial according to Holland's theory, such as the \textit{Investigative} personality. While personality assignments can enhance the ability to linguistically simulate roles, their impact on performance in domain-specific QA tasks remains inconclusive and may be influenced by other factors. This suggests the need for further investigation into how cognitive and sociolinguistic perspectives could inform the application of LLMs to complex tasks.

\subsection{Evaluation of Profession-Based Persona Performance}\label{sec:result_PBP}
Table \ref{tab:profession} shows changes in QA performance with one OPBP and five different PBPs: \textit{Scientific Person}, \textit{Scientist}, \textit{Biologist}, \textit{Chemist}, and \textit{Physicist}, along with \textit{Artist} as a negative control, based on the observations from Figure \ref{fig:personality}. \textit{Scientific Person} closely resembles \textit{Scientist} in wording, yet it aligns more with cognitive styles rather than direct domain knowledge, making it better classified as an OPBP rather than a PBP. The \textit{Scientist} persona generally led to higher QA accuracy compared to the \textit{Scientific Person}, particularly in cases where the baseline score was low and the task required substantial domain knowledge. This demonstrates that PBPs, which are directly associated with experts, are more effective than OPBPs in invoking domain knowledge. 

A more targeted assignment of PBPs aligned with specific domains also led to an improvement in domain knowledge utilization. For example, in GPT-3.5, the \textit{Biologist} persona improved performance by 0.7\%, and the \textit{Chemist} increased scores by 2.22\%. Similarly, in Llama-3-8B, the \textit{Biologist} improved performance by 3.09\%, and the \textit{Chemist} by 2.0\%. Along with noticeable performance changes, the \textit{Biologist} and \textit{Chemist} exhibited relatively higher CR and AR when applied to less relevant domains compared to the \textit{Physicist}. This implies that physics-related tasks may exhibit different characteristics from other domains, a point further explored in Section \ref{sec:limitation}.

On the other hand, when the model was assigned a persona that was misaligned with the task, performance often dropped significantly. In GPT-3.5, the \textit{Artist} persona experienced a significant 23.96\% drop in performance compared to the \textit{Scientist} persona on the College Physics benchmark, exhibiting the highest CR and AR across all datasets. This aligns with the expectation that the \textit{Artist} persona limits the model's ability to explore and apply knowledge in scientific domains. Interestingly, a notable 11.43\% performance gap was observed between the PBP (\textit{Artist}) and the OPBP (\textit{Artistic Person}) on the same benchmark in GPT-3.5 (refer to Appendix \ref{app:gpt_results}). Although both personas share the same word, ``artist,'' as the input prompt, this large performance gap suggests that the model could recognize distinctions between professional roles and personality-based identities to some extent, a pattern also observed in the comparison between the \textit{Scientific Person} and \textit{Scientist} personas. In Llama-3-8B, \textit{Artist} persona causes a smaller performance difference compared to GPT-3.5. This suggests that the effect of personas can quantitatively vary across different LLM architectures, which we further examine in Appendix \ref{app:gpt_results} and \ref{app:llama_results}.

\section{Do LLMs Differentiate Personas?}\label{neuron_Results}
In the previous sections, we observed performance variations across fine-grained persona assignments. In this section, we explore potential underlying reasons for these differences by analyzing the model’s internal mechanisms. To understand how LLMs retrieve and utilize domain-specific knowledge, we adopt the perspective that feed-forward layers function as key-value memory structures responsible for storing and recalling knowledge \cite{FFN_key_value_21}. We assess how the model's knowledge invocation process varies across different persona conditions by examining changes in activated parameters using neuron detection \cite{knowledge_neuron}. Here, neurons are defined as individual rows or columns within the parameter matrices of a language model, and we adopt the neuron detection methodology proposed in \cite{Multilingualism}:
\begin{align}
    \text{NeuronImp}(W_{\text{up}}[:, k] | c) = \left\| \hat{\text{FFN}}(x) - \text{FFN}(x) \right\|2 \notag \\
    = \left\| \left( h_{\text{ffn}} \cdot \text{Mask}[k] \right) W_{\text{down}}(x) \right\|_2
\end{align}

To investigate how persona assignments influence question answering, we analyze neuron activations based on input \(c\), which consists of both the persona description and the question data. The importance of the \(k\)-th neuron (NeuronImp) in \(W_{up}\) is computed based on \(c\), where \(x\) represents the embedding processed by the feed-forward network (FFN). The NeuronImp is defined as the magnitude of change in the FFN output when the \(k\)-th neuron is deactivated. Typically, neuron specificity for a given input is determined by whether the computed importance exceeds a predefined threshold. However, rather than applying a threshold-based criterion, we report the raw NeuronImp values directly to analyze the degree of importance variation across different personas. This approach allows us to examine neuron activation shifts and assess their relative contribution to the inference process under distinct persona assignments.

\begin{figure}[t!]
    \centering    
    \includegraphics[width=1\linewidth]{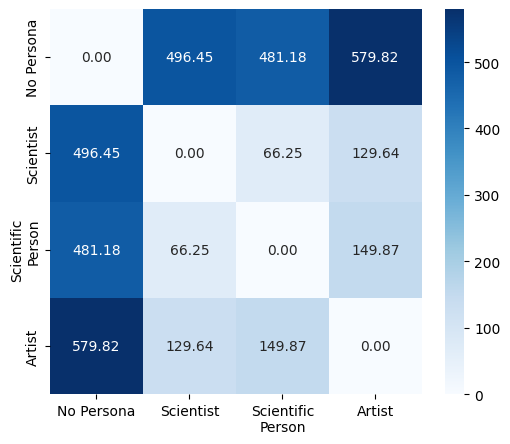}
    \caption{Pairwise Euclidean distance of neuron impact across different personas in the Llama-3-8B model, showing how the model's internal activations vary based on persona assignment.}
    \label{fig:similar}
\end{figure}

\subsection{Similarity and Differences Across Personas}
Figure \ref{fig:similar} presents the pairwise Euclidean distance calculated for NeuronImp across different input texts, illustrating how the model responds differently depending on the assigned persona. The values represent the degree of NeuronImp variation between different persona settings, with higher values indicating greater differences in how the model processes information.

A key observation is that the no-persona condition exhibits the largest differences compared to any persona, particularly with the \textit{Artist} persona. This could suggest that the \textit{Artist} persona induces activation patterns that are more distinct from the baseline, possibly due to weaker alignment with task-specific knowledge retrieval.

The \textit{Scientist} and \textit{Scientific Person} personas are both recognized as relevant to scientific contexts. However, the slight variation between the two raises the possibility that persona specificity may still influence model activations to some degree. These findings point to a potential role of persona conditions in shaping how knowledge is retrieved and utilized, emphasizing the need to consider factors beyond mere relevance and fundamental meaning in future persona applications.

\subsection{Layer-Wise Impact of Personas}
\begin{figure}[t!]
    \centering    
    \includegraphics[width=1\linewidth]{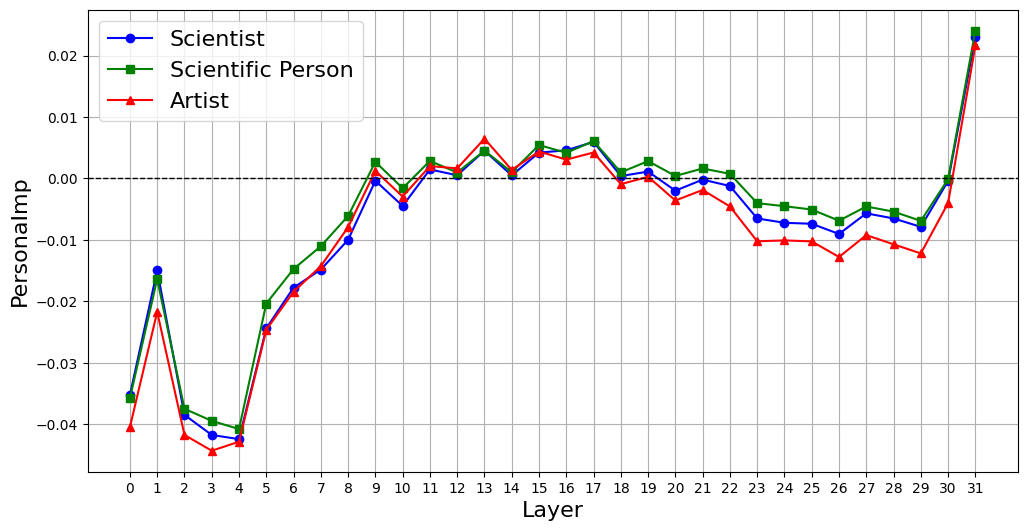}
    \caption{Layer-wise neuron impact differences between the no-persona condition and various persona assignments (\textit{Scientist}, \textit{Scientific Person}, \textit{Artist}) in the Llama-3-8B model.}
    \label{fig:layer}
\end{figure}


To compare how strongly neurons contribute to model inference at different layers under different persona conditions, we define the total contribution of all neurons in layer \(l\) as:
\begin{equation}
   \text{Contrib}(l) = \sum_{k=1}^{N_{l}} \text{NeuronImp}_{k,l}
\end{equation}

where Contrib(\(l\)) represents the aggregate NeuronImp in layer \(l\) and \(N_{l}\) is the total number of neurons in that layer. Based on this, the persona-induced impact (PersonaImp) at each layer is quantified as follows:

\begin{align}
   \text{PersonaImp}(l)_{\text{Persona}}
   = \log(\text{Contrib}(l)_{\text{Persona}}) \notag \\ - \log(\text{Contrib}(l)_{\text{No Persona}})
\end{align}

Figure \ref{fig:layer} shows the variation in total neuron impact across layers between the no-persona condition and different persona assignments. The graph tracks how neuron activation fluctuates across layers, providing insight into where persona influences become more pronounced.

In earlier layers, which are typically associated with lexical or syntactic processing and query understanding \cite{dola}, persona assignments appear to reduce overall neuron engagement. This may suggest that persona conditioning helps minimize ambiguity in token interpretation, leading to a reduced need for additional processing at this stage. However, as processing progresses into the later layers, a noticeable divergence emerges, particularly in the final layer. This shift suggests that persona assignments may play a role in shaping the model’s decision-making and response formulation in the final stages. While further investigation is required, this pattern suggests that personas could have a greater influence on high-level reasoning and knowledge recall, rather than fundamental linguistic processing.

\section{Related Work}
Persona adoption has been extensively studied in dialogue systems, including the use of LLMs to emulate various characters \cite{park2023generative}, and ensuring consistent personas in user interactions or character simulations \cite{li2023chatharuhi, gao-etal-2023-peacok}. Research has explored various types of personas, covering socio-demographic characteristics such as race, gender, and profession \cite{parrot_2023}, personality types \cite{jiang2024personallm}, and fictional or real-life figures \cite{2023-character}. These studies have also addressed ethical issues related to toxicity \cite{toxicity_2023} and bias \cite{sheng2021revealing} in text generated by certain personas. Some studies have introduced personas in QA for reasoning tasks \cite{zero_shot_role, bias_2024}

The potential of personas to enhance domain knowledge remains underexamined. A previous study found that while in-domain personas led to better performance than out-of-domain personas, the actual performance gains were minimal \cite{helpful_assistent}. However, the distinct characteristics of different roles and their impact on model internals have not been sufficiently analyzed beyond surface-level similarity. Our study distinguishes between personas directly linked to domain expertise and those with only indirect relevance, allowing for a more fine-grained examination of how LLMs process and respond to different persona types in domain-specific tasks.


\section{Conclusion}
In this study, we explored whether personas enhance performance in domain-specific tasks and how LLMs interpret professions and personality traits. We designed two types of personas: Profession-Based Personas (PBPs), which explicitly represent domain expertise, and Occupational Personality-Based Personas (OPBPs), which reflect cognitive tendencies rather than professional knowledge. Using tailored prompts and metrics, we analyzed their impact on domain-specific QA tasks. 

Our findings indicate that LLMs respond differently to various persona traits and can adapt to each assigned role's unique characteristics. PBPs, closely tied to domain knowledge, showed potential to enhance performance. In contrast, OPBPs often led to decreased performance, even when they were task-relevant. This suggests that linguistic role-play alone does not guarantee effective knowledge utilization and that some personas may introduce biases that hinder reasoning.

To further investigate whether persona differentiation occurs beyond linguistic adaptation, we analyzed neuron activation patterns. Our findings suggest that the effects of personas become more pronounced in high-level knowledge retrieval processes. Additionally, our analysis suggests that LLMs may differentiate professional roles from occupational traits to some extent, indicating a possible tendency toward internal persona differentiation consistent with human understanding. These findings highlight the need for further research on how LLMs encode and utilize role information.

\section{Discussion}\label{sec:limitation} 
\textbf{Expanding the Scope of Analysis.} Our study explored the impact of occupational personas, categorizing them into professional and personality-based types based on Holland's six categories. While these categories provide a structured approach to personality traits, they do not generalize to all possible occupational types. We proposed approaches to evaluate the impact of occupational personas on performance, with experiments primarily focused on science domain datasets, analyzing PBPs such as biologists, chemists, and physicists. However, a broader range of occupational personas can be applied to various domains and tasks. Future studies could expand on this by applying professional and personality-based personas in various fields, using a wider range of LLMs, to explore the applicability and effectiveness of our findings across broader contexts.

\textbf{Persona-Based Prompting Approach.} Although persona-based prompting can improve LLM behavior, our findings indicate that assigning personas must be approached with caution. We acknowledge that alternative methods such as prompt tuning or fine-tuning the LLMs on domain-specific data are also considered to enhance specialized tasks. These methods, however, involve huge computational resources and may not promote a human-like understanding of LLMs as effectively as persona-based approach. This is why we proposed the occupational persona-based approach to apply a practical and accessible method for specific domain QA tasks, by simulating human-like manner.

\begin{table}[t!]
\centering
\begin{tabular}{l|c|c|c}
    \hline
             & \textbf{Biologist} & \textbf{Chemist} & \textbf{Physicist} \\
    \hline
    \textbf{Biology}    & -      & -3.01   & -3.32   \\
    \textbf{Chemistry}  & -5.79  & -       & 3.11    \\
    \textbf{Physics}    & -20.48 & -15.95  & -       \\
    \hline
\end{tabular}
\caption{Performance differences across domains for profession personas in GPT-3.5.}
\label{tab:relative_change}
\end{table}

\textbf{Task-Specific Trends.} In our analysis, we observed an unanticipated trend in the College Physics dataset and with the Physicist persona, differing from other domains. Notably, in the GPT-3.5 results on the physics dataset, none of the personas outperformed the No Persona. To better understand how the model perceives these three major scientific domains, we conducted experiments (Section \ref{sec:result_PBP}) by substituting domain-specific personas with alternative personas. For instance, using a Biologist for the College Biology dataset represents an appropriate pairing, whereas using a Physicist for the same dataset serves as an alternative pairing. 

As previously mentioned, the Biologist and Chemist exhibited relatively high CR and AR when applied to other domains, which is also reflected in the accuracy results. Table \ref{tab:relative_change} presents the relative performance change when comparing domain-specific personas with alternative personas. The most significant changes were observed in the Physics dataset. This difference indicates that the application of personas in physics-related tasks does not facilitate the same performance improvements observed in biology and chemistry. One possible explanation is that physics often involves abstract concepts and mathematical reasoning that may not be as effectively captured through persona-based prompts. LLMs might have a different internal representation or understanding of physics compared to other sciences in terms of reasoning, making it less responsive to persona assignments. Therefore, Physics may require different prompting strategies or additional contextual information to improve performance due to their quantitative problem-solving and abstract principles.

\textbf{Challenges in Distinguishing Between Reasoning and Knowledge-Intensive Tasks.} Some studies suggest that prompts designed for reasoning tasks do not necessarily improve performance on knowledge-intensive tasks \cite{CMMMLU}. Given that MMLU tasks require both knowledge and reasoning ability, it would be beneficial to explore a systematic classification of tasks into reasoning and domain knowledge categories. Further research is also needed to explore how personas affect performance across these different task types.




\begin{thebibliography}{25}
\providecommand{\natexlab}[1]{#1}

\bibitem[{Chuang et~al.(2024)Chuang, Xie, Luo, Kim, Glass, and He}]{dola}
Yung-Sung Chuang, Yujia Xie, Hongyin Luo, Yoon Kim, James~R. Glass, and
  Pengcheng He. 2024.
\newblock \href {https://openreview.net/forum?id=Th6NyL07na} {Dola: Decoding by
  contrasting layers improves factuality in large language models}.
\newblock In \emph{The Twelfth International Conference on Learning
  Representations}.

\bibitem[{Dai et~al.(2022)Dai, Dong, Hao, Sui, Chang, and
  Wei}]{knowledge_neuron}
Damai Dai, Li~Dong, Yaru Hao, Zhifang Sui, Baobao Chang, and Furu Wei. 2022.
\newblock \href {https://doi.org/10.18653/v1/2022.acl-long.581} {Knowledge
  neurons in pretrained transformers}.
\newblock In \emph{Proceedings of the 60th Annual Meeting of the Association
  for Computational Linguistics (Volume 1: Long Papers)}, pages 8493--8502,
  Dublin, Ireland. Association for Computational Linguistics.

\bibitem[{Deshpande et~al.(2023)Deshpande, Murahari, Rajpurohit, Kalyan, and
  Narasimhan}]{toxicity_2023}
Ameet Deshpande, Vishvak Murahari, Tanmay Rajpurohit, Ashwin Kalyan, and
  Karthik Narasimhan. 2023.
\newblock \href {https://doi.org/10.18653/v1/2023.findings-emnlp.88} {Toxicity
  in chatgpt: Analyzing persona-assigned language models}.
\newblock In \emph{Findings of the Association for Computational Linguistics:
  EMNLP 2023}, pages 1236--1270, Singapore. Association for Computational
  Linguistics.

\bibitem[{Gao et~al.(2023)Gao, Borges, Oh, Bayazit, Kanno, Wakaki, Mitsufuji,
  and Bosselut}]{gao-etal-2023-peacok}
Silin Gao, Beatriz Borges, Soyoung Oh, Deniz Bayazit, Saya Kanno, Hiromi
  Wakaki, Yuki Mitsufuji, and Antoine Bosselut. 2023.
\newblock \href {https://doi.org/10.18653/v1/2023.acl-long.362} {{P}ea{C}o{K}:
  Persona commonsense knowledge for consistent and engaging narratives}.
\newblock In \emph{Proceedings of the 61st Annual Meeting of the Association
  for Computational Linguistics (Volume 1: Long Papers)}, pages 6569--6591,
  Toronto, Canada. Association for Computational Linguistics.

\bibitem[{Geva et~al.(2021)Geva, Schuster, Berant, and Levy}]{FFN_key_value_21}
Mor Geva, Roei Schuster, Jonathan Berant, and Omer Levy. 2021.
\newblock \href {https://doi.org/10.18653/v1/2021.emnlp-main.446} {Transformer
  feed-forward layers are key-value memories}.
\newblock In \emph{Proceedings of the 2021 Conference on Empirical Methods in
  Natural Language Processing}, pages 5484--5495, Online and Punta Cana,
  Dominican Republic. Association for Computational Linguistics.

\bibitem[{Gupta et~al.(2024)Gupta, Shrivastava, Deshpande, Kalyan, Clark,
  Sabharwal, and Khot}]{bias_2024}
Shashank Gupta, Vaishnavi Shrivastava, Ameet Deshpande, Ashwin Kalyan, Peter
  Clark, Ashish Sabharwal, and Tushar Khot. 2024.
\newblock \href {https://openreview.net/forum?id=kGteeZ18Ir} {Bias runs deep:
  Implicit reasoning biases in persona-assigned {LLM}s}.
\newblock In \emph{The Twelfth International Conference on Learning
  Representations}.

\bibitem[{Hendrycks et~al.(2021)Hendrycks, Burns, Basart, Zou, Mazeika, Song,
  and Steinhardt}]{MMLU}
Dan Hendrycks, Collin Burns, Steven Basart, Andy Zou, Mantas Mazeika, Dawn
  Song, and Jacob Steinhardt. 2021.
\newblock \href {https://openreview.net/forum?id=d7KBjmI3GmQ} {Measuring
  massive multitask language understanding}.
\newblock In \emph{International Conference on Learning Representations}.

\bibitem[{Holland(1985)}]{holland1985}
J.L. Holland. 1985.
\newblock \href {https://books.google.co.kr/books?id=8QxBAAAAMAAJ}
  {\emph{Making Vocational Choices: A Theory of Vocational Personalities and
  Work Environments}}.
\newblock Prentice-Hall series in counseling and human development.
  Prentice-Hall.

\bibitem[{Jiang et~al.(2024)Jiang, Zhang, Cao, Breazeal, Roy, and
  Kabbara}]{jiang2024personallm}
Hang Jiang, Xiajie Zhang, Xubo Cao, Cynthia Breazeal, Deb Roy, and Jad Kabbara.
  2024.
\newblock \href {https://arxiv.org/abs/2305.02547} {Personallm: Investigating
  the ability of large language models to express personality traits}.
\newblock \emph{Preprint}, arXiv:2305.02547.

\bibitem[{Kong et~al.(2024)Kong, Zhao, Chen, Li, Qin, Sun, Zhou, Wang, and
  Dong}]{zero_shot_role}
Aobo Kong, Shiwan Zhao, Hao Chen, Qicheng Li, Yong Qin, Ruiqi Sun, Xin Zhou,
  Enzhi Wang, and Xiaohang Dong. 2024.
\newblock \href {https://arxiv.org/abs/2308.07702} {Better zero-shot reasoning
  with role-play prompting}.
\newblock \emph{Preprint}, arXiv:2308.07702.

\bibitem[{Li et~al.(2023)Li, Leng, Yan, Shen, Wang, MI, Fei, Feng, Yan, Wang,
  Zhan, Jia, Wu, and Sun}]{li2023chatharuhi}
Cheng Li, Ziang Leng, Chenxi Yan, Junyi Shen, Hao Wang, Weishi MI, Yaying Fei,
  Xiaoyang Feng, Song Yan, HaoSheng Wang, Linkang Zhan, Yaokai Jia, Pingyu Wu,
  and Haozhen Sun. 2023.
\newblock \href {https://arxiv.org/abs/2308.09597} {Chatharuhi: Reviving anime
  character in reality via large language model}.
\newblock \emph{Preprint}, arXiv:2308.09597.

\bibitem[{Li et~al.(2024)Li, Zhang, Koto, Yang, hai zhao, Gong, Duan, and
  Baldwin}]{CMMMLU}
Haonan Li, Yixuan Zhang, Fajri Koto, Yifei Yang, hai zhao, Yeyun Gong, Nan
  Duan, and Timothy Baldwin. 2024.
\newblock \href {https://openreview.net/forum?id=ck4SG9lnrQ} {{CMMLU}:
  Measuring massive multitask language understanding in chinese}.

\bibitem[{Oh et~al.(2023)Oh, Lee, Li, and Wang}]{persona_knowledge_retrieval}
Minsik Oh, Joosung Lee, Jiwei Li, and Guoyin Wang. 2023.
\newblock \href {https://doi.org/10.18653/v1/2023.emnlp-main.1020} {{PK}-{ICR}:
  Persona-knowledge interactive multi-context retrieval for grounded dialogue}.
\newblock In \emph{Proceedings of the 2023 Conference on Empirical Methods in
  Natural Language Processing}, pages 16383--16395, Singapore. Association for
  Computational Linguistics.

\bibitem[{OpenAI(2022)}]{openai}
OpenAI. 2022.
\newblock \href {https://openai.com/index/chatgpt/} {Introducing chatgpt}.
\newblock Accessed: 2024-09-15.

\bibitem[{Park et~al.(2023)Park, O'Brien, Cai, Morris, Liang, and
  Bernstein}]{park2023generative}
Joon~Sung Park, Joseph~C. O'Brien, Carrie~J. Cai, Meredith~Ringel Morris, Percy
  Liang, and Michael~S. Bernstein. 2023.
\newblock \href {https://arxiv.org/abs/2304.03442} {Generative agents:
  Interactive simulacra of human behavior}.
\newblock \emph{Preprint}, arXiv:2304.03442.

\bibitem[{Qin(2024)}]{quantization}
Minghai Qin. 2024.
\newblock \href {https://arxiv.org/abs/2408.15301} {The uniqueness of
  llama3-70b with per-channel quantization: An empirical study}.
\newblock \emph{Preprint}, arXiv:2408.15301.

\bibitem[{Shao et~al.(2023)Shao, Li, Dai, and Qiu}]{2023-character}
Yunfan Shao, Linyang Li, Junqi Dai, and Xipeng Qiu. 2023.
\newblock \href {https://aclanthology.org/2023.emnlp-main.814/}
  {Character-{LLM}: A trainable agent for role-playing}.
\newblock In \emph{Proceedings of the 2023 Conference on Empirical Methods in
  Natural Language Processing}, pages 13153--13187, Singapore. Association for
  Computational Linguistics.

\bibitem[{Sheng et~al.(2021)Sheng, Arnold, Yu, Chang, and
  Peng}]{sheng2021revealing}
Emily Sheng, Josh Arnold, Zhou Yu, Kai-Wei Chang, and Nanyun Peng. 2021.
\newblock \href {https://arxiv.org/abs/2104.08728} {Revealing persona biases in
  dialogue systems}.
\newblock \emph{Preprint}, arXiv:2104.08728.

\bibitem[{Tseng et~al.(2024)Tseng, Huang, Hsiao, Chen, Huang, Meng, and
  Chen}]{persona_survey_recent}
Yu-Min Tseng, Yu-Chao Huang, Teng-Yun Hsiao, Wei-Lin Chen, Chao-Wei Huang,
  Yu~Meng, and Yun-Nung Chen. 2024.
\newblock \href {https://doi.org/10.18653/v1/2024.findings-emnlp.969} {Two
  tales of persona in {LLM}s: A survey of role-playing and personalization}.
\newblock In \emph{Findings of the Association for Computational Linguistics:
  EMNLP 2024}, pages 16612--16631, Miami, Florida, USA. Association for
  Computational Linguistics.

\bibitem[{Wan et~al.(2023)Wan, Zhao, Chadha, Peng, and Chang}]{parrot_2023}
Yixin Wan, Jieyu Zhao, Aman Chadha, Nanyun Peng, and Kai-Wei Chang. 2023.
\newblock \href {https://doi.org/10.18653/v1/2023.findings-emnlp.648} {Are
  personalized stochastic parrots more dangerous? evaluating persona biases in
  dialogue systems}.
\newblock In \emph{Findings of the Association for Computational Linguistics:
  EMNLP 2023}, pages 9677--9705, Singapore. Association for Computational
  Linguistics.

\bibitem[{Wang et~al.(2024{\natexlab{a}})Wang, Xiao, tse Huang, Yuan, Xu, Guo,
  Tu, Fei, Leng, Wang, Chen, Li, and Xiao}]{wang2024incharacter}
Xintao Wang, Yunze Xiao, Jen tse Huang, Siyu Yuan, Rui Xu, Haoran Guo, Quan Tu,
  Yaying Fei, Ziang Leng, Wei Wang, Jiangjie Chen, Cheng Li, and Yanghua Xiao.
  2024{\natexlab{a}}.
\newblock \href {https://arxiv.org/abs/2310.17976} {Incharacter: Evaluating
  personality fidelity in role-playing agents through psychological
  interviews}.
\newblock \emph{Preprint}, arXiv:2310.17976.

\bibitem[{Wang et~al.(2024{\natexlab{b}})Wang, zhongyuan peng, Que, Liu, Zhou,
  Wu, Guo, Gan, Ni, Zhang, Zhang, Ouyang, Xu, Chen, Fu, and
  Peng}]{wang2024rolellm}
Zekun Wang, zhongyuan peng, Haoran Que, Jiaheng Liu, Wangchunshu Zhou, Yuhan
  Wu, Hongcheng Guo, Ruitong Gan, Zehao Ni, Man Zhang, Zhaoxiang Zhang, Wanli
  Ouyang, Ke~Xu, Wenhu Chen, Jie Fu, and Junran Peng. 2024{\natexlab{b}}.
\newblock \href {https://openreview.net/forum?id=i4ULDEeBss} {Role{LLM}:
  Benchmarking, eliciting, and enhancing role-playing abilities of large
  language models}.

\bibitem[{Wang et~al.(2024{\natexlab{c}})Wang, Mao, Wu, Ge, Wei, and
  Ji}]{unleash_persona_factual}
Zhenhailong Wang, Shaoguang Mao, Wenshan Wu, Tao Ge, Furu Wei, and Heng Ji.
  2024{\natexlab{c}}.
\newblock \href {https://arxiv.org/abs/2307.05300} {Unleashing the emergent
  cognitive synergy in large language models: A task-solving agent through
  multi-persona self-collaboration}.
\newblock \emph{Preprint}, arXiv:2307.05300.

\bibitem[{Zhao et~al.(2024)Zhao, Zhang, Chen, Kawaguchi, and
  Bing}]{Multilingualism}
Yiran Zhao, Wenxuan Zhang, Guizhen Chen, Kenji Kawaguchi, and Lidong Bing.
  2024.
\newblock \href {https://arxiv.org/abs/2402.18815} {How do large language
  models handle multilingualism?}
\newblock \emph{Preprint}, arXiv:2402.18815.

\bibitem[{Zheng et~al.(2024)Zheng, Pei, Logeswaran, Lee, and
  Jurgens}]{helpful_assistent}
Mingqian Zheng, Jiaxin Pei, Lajanugen Logeswaran, Moontae Lee, and David
  Jurgens. 2024.
\newblock \href {https://doi.org/10.18653/v1/2024.findings-emnlp.888} {When
  {\textquotedblright}a helpful assistant{\textquotedblright} is not really
  helpful: Personas in system prompts do not improve performances of large
  language models}.
\newblock In \emph{Findings of the Association for Computational Linguistics:
  EMNLP 2024}, pages 15126--15154, Miami, Florida, USA. Association for
  Computational Linguistics.

\end{thebibliography}

\appendix\label{sec:appendix} 

\section{Overall Experimental Results}
In the overall experiment, we used a baseline with no persona and assigned 18 different personas to three LLMs across three datasets in the MMLU benchmark: college biology, college chemistry, and college physics. The 18 personas included five PBPs, seven OPBPs-Brief, and six OPBPs-Descriptive. The performance results tables show the average accuracy, average CR, and average AR of a total of nine runs, with three runs for each persona, using three different persona instructions. Each row of the table represents one of the 18 different personas, and the columns denote accuracy, CR, and AR for each dataset. The highest performance for each dataset is indicated in bold.

\subsection{GPT-3.5-Turbo \& GPT-4o-mini}\label{app:gpt_results}
Table \ref{tab:gpt3.5} present the results for GPT-3.5-Turbo \cite{openai} and Table \ref{tab:gpt4o} for GPT-4o mini. We configured a maximum token length of 1024, a temperature of 0, and a top-p value of 1.

GPT-3.5 shows higher CR and AR when the model perceives a mismatch between the task and persona. In contrast, GPT-4o mini has a higher CR for the artist and artistic personas, but AR remains close to zero. This could suggest that the newer model better distinguishes between personality and knowledge. However, despite the low AR, the occasional performance drop for the Artist persona indicates that there may still be inherent limitations in how the model handles domain knowledge and the importance of relevant personas.


\begin{table*}[t!]
\centering
\begin{tabular}{l|ccc|ccc|ccc}
\hline
                            & \multicolumn{3}{c|}{\textbf{College Biology}} & \multicolumn{3}{c|}{\textbf{College Chemistry}} & \multicolumn{3}{c}{\textbf{College Physics}} \\ \hline
\textbf{Persona }                   & \textbf{Acc}   & \textbf{CR}  & \textbf{AR}  & \textbf{Acc}    & \textbf{CR}     & \textbf{AR}     & \textbf{Acc}   & \textbf{CR}   & \textbf{AR}     \\ \hline
\textbf{No Persona }                 & 73.61            & -       & -       & 47.67             & -        & -       & \textbf{54.25}   & -       & -      \\
\textbf{Scientist }                  & 74.23            & 1.46    & 0.15    & \textbf{51.56}    & 1.89     & 0.44    & 52.18            & 1.42    & 0.77   \\
\textbf{Biologist   }                & \textbf{74.31}   & 6.02    & 0.31    & 47.00             & 26.44    & 13.78   & 40.19            & 43.68   & 22.44  \\
\textbf{Chemist  }                   & 72.07            & 11.04   & 2.08    & 49.89             & 1.78     & 1.00    & 42.48            & 11.11   & 4.03   \\
\textbf{Physicist }                  & 71.84            & 1.77    & 0.69    & 51.44             & 0.00     & 0.00    & 50.54            & 0.00    & 0.00   \\
\textbf{Artist  }                    & 58.64            & 66.90   & 16.74   & 30.56             & 68.44    & 39.67   & 28.22            & 66.67   & 40.41  \\ \hline
\textbf{Scientific (Brief)  }        & 73.76            & 11.27   & 1.08    & 45.89             & 3.67     & 1.00    & 50.33            & 4.25    & 0.54   \\
\textbf{Realistic (Brief)}           & 69.60            & 35.96   & 6.48    & 36.67             & 49.67    & 31.22   & 39.43            & 56.10   & 28.00  \\
\textbf{Investigative (Brief)  }     & 71.76            & 3.17    & 0.39    & 46.44             & 3.33     & 1.00    & 48.04            & 1.96    & 1.31   \\
\textbf{Artistic (Brief) }           & 65.20            & 57.48   & 6.87    & 37.78             & 65.78    & 24.22   & 39.65            & 63.51   & 16.01  \\
\textbf{Social (Brief)   }           & 70.37            & 20.99   & 6.48    & 43.56             & 16.67    & 31.22   & 46.62            & 13.62   & 28.00  \\
\textbf{Enterprising (Brief) }       & 72.76            & 34.72   & 1.47    & 43.33             & 28.67    & 3.44    & 50.11            & 25.82   & 2.61   \\
\textbf{Conventional (Brief) }       & 71.68            & 26.93   & 2.47    & 42.67             & 25.67    & 9.44    & 42.48            & 18.52   & 5.23   \\ \hline
\textbf{Realistic (Descriptive) }    & 69.37            & 57.25   & 4.48    & 40.78             & 86.67    & 27.67   & 47.71            & 78.32   & 16.12  \\
\textbf{Investigative (Descriptive)} & 70.37            & 33.18   & 1.93    & 46.33             & 31.89    & 3.89    & 51.85            & 18.84   & 0.98   \\
\textbf{Artistic (Descriptive)  }    & 57.02            & 91.82   & 18.29   & 32.11             & 96.67    & 50.00   & 33.88            & 88.78   & 33.55  \\
\textbf{Social (Descriptive) }       & 67.05            & 88.58   & 7.02    & 35.89             & 98.33    & 41.33   & 32.25            & 96.08   & 35.84  \\
\textbf{Enterprising (Descriptive)}  & 64.43            & 76.77   & 8.87    & 39.11             & 93.44    & 33.11   & 37.47            & 86.60   & 23.86  \\
\textbf{Conventional (Descriptive)}  & 71.37            & 41.98   & 1.70    & 46.55             & 48.22    & 7.44    & 48.69            & 33.88   & 2.72        \\  \hline
\end{tabular}
    \caption{Results of GPT-3.5-Turbo.}
    \label{tab:gpt3.5}
\end{table*}

\begin{table*}[t!]
\centering
\begin{tabular}{l|ccc|ccc|ccc}
\hline
                            & \multicolumn{3}{c|}{\textbf{College Biology}} & \multicolumn{3}{c|}{\textbf{College Chemistry}} & \multicolumn{3}{c}{\textbf{College Physics}} \\ \hline
\textbf{Persona}                    & \textbf{Acc}   & \textbf{CR}  & \textbf{AR}  & \textbf{Acc}    & \textbf{CR}     & \textbf{AR}     & \textbf{Acc}   & \textbf{CR}   & \textbf{AR}     \\ \hline
\textbf{No Persona}                 & \textbf{95.14}    & -       & -      & 63.33             & -        & -       & \textbf{89.87}   & -       & -      \\
\textbf{Scientist}                  & 93.98             & 0.00    & 0.00   & 61.55             & 0.00     & 0.00    & 87.04            & 0.00    & 0.00   \\
\textbf{Biologist}                   & 94.44             & 0.00    & 0.00   & 64.22             & 0.33     & 0.33    & 88.78            & 0.22    & 0.00   \\
\textbf{Chemist}                     & 93.44             & 0.23    & 0.00   & 63.22             & 0.00     & 0.00    & 89.54            & 0.00    & 0.00   \\
\textbf{Physicist}                   & 93.98             & 0.00    & 0.00   & 62.78             & 0.00     & 0.00    & 88.45            & 0.00    & 0.00   \\
\textbf{Artist}                      & 92.52             & 40.35   & 0.00   & 62.78             & 30.22    & 0.11    & 85.19            & 27.12   & 0.00   \\ \hline
\textbf{Scientific (Brief)}          & 93.75             & 0.00    & 0.00   & 63.22             & 0.00     & 0.00    & 87.91            & 0.00    & 0.00   \\
\textbf{Realistic (Brief)}           & 93.52             & 0.00    & 0.00   & 63.00             & 0.00     & 0.00    & 85.84            & 0.00    & 0.00   \\
\textbf{Investigative (Brief)}       & 93.06             & 0.00    & 0.00   & 62.11             & 0.00     & 0.00    & 85.84            & 0.00    & 0.00   \\
\textbf{Artistic (Brief)}            & 91.05             & 0.00    & 0.00   & 62.56             & 0.00     & 0.00    & 85.51            & 0.00    & 0.00   \\
\textbf{Social (Brief)}              & 93.75             & 0.00    & 0.00   & \textbf{64.45}    & 0.00     & 0.00    & 84.42            & 0.00    & 0.00   \\
\textbf{Enterprising (Brief)}        & 93.83             & 0.00    & 0.00   & 61.78             & 0.00     & 0.00    & 87.04            & 0.00    & 0.00   \\
\textbf{Conventional (Brief)}        & 93.36             & 0.00    & 0.00   & 62.22             & 0.00     & 0.00    & 86.17            & 0.00    & 0.00   \\ \hline
\textbf{Realistic (Descriptive)}     & 93.13             & 0.00    & 0.00   & 61.78             & 0.00     & 0.00    & 84.64            & 0.00    & 0.00   \\
\textbf{Investigative (Descriptive)} & 94.68             & 0.00    & 0.00   & 62.56             & 0.00     & 0.00    & 87.14            & 0.00    & 0.00   \\
\textbf{Artistic (Descriptive)}      & 93.06             & 32.02   & 0.00   & 62.78             & 16.44    & 0.11    & 84.86            & 9.69    & 0.00   \\
\textbf{Social (Descriptive)}        & 94.52             & 1.54    & 0.00   & 62.11             & 7.11     & 0.44    & 83.55            & 3.38    & 0.00   \\
\textbf{Enterprising (Descriptive)}  & 93.83             & 9.33    & 0.00   & 62.89             & 4.33     & 0.00    & 83.99            & 2.61    & 0.00   \\
\textbf{Conventional (Descriptive)}  & 94.37             & 0.00    & 0.00   & 63.22             & 0.00     & 0.00    & 83.66            & 0.00    & 0.00        \\  \hline
\end{tabular}
    \caption{Results of GPT-4o mini.}
    \label{tab:gpt4o}
\end{table*}

\subsection{Llama-3-8B-Instruct \& Llama-3-70B-Instruct}\label{app:llama_results}
Additionally, our experiments included recent models by Meta, such as Meta-Llama-3-8B-Instruct and Meta-Llama-3-70B-Instruct \footnote{\url{https://llama.meta.com/llama3/}}, utilizing 8-bit quantization with bitsandbytes, Transformers by Huggingface\footnote{\url{https://github.com/TimDettmers/bitsandbytes}}. For both Llama-3 models, we configured a maximum number of new tokens to be 1024, set 'do\_sample' to 'True', the temperature to 'None', and the top-p value to 1. Table \ref{tab:8b} and Table \ref{tab:70b} present the results for the Llama models.

Previous studies have noted significant performance drops in Llama-70B with Per-Channel Quantization \cite{quantization}. As a result, the baseline performance for the 70B model is lower than the 8B model. However, for this study, we chose to focus on the influence of personas rather than the final performance of the models.

Compared to GPT, performance changes in Llama models are less pronounced, and they exhibit stronger robustness to personality prompts. The trend that PBPs are more closely tied to domain knowledge than OPBPs remains consistent. 

In the 8B model, CR is generally higher, with the model often responding with statements like ``As a [persona], I will answer,'' rather than admitting its limitations. ARs are notably low, with negligible instances of the model completely refusing to answer. Explicit persona application is less frequent in responses compared to GPT, with less variation in performance. Interestingly, there are more cases where the 8B model's performance improves with personas than in GPT.



\begin{table*}[t!]
\centering
\begin{tabular}{l|ccc|ccc|ccc}
\hline
                            & \multicolumn{3}{c|}{\textbf{College Biology}} & \multicolumn{3}{c|}{\textbf{College Chemistry}} & \multicolumn{3}{c}{\textbf{College Physics}} \\ \hline
\textbf{Persona}                    & \textbf{Acc}   & \textbf{CR}  & \textbf{AR}  & \textbf{Acc}    & \textbf{CR}     & \textbf{AR}     & \textbf{Acc}   & \textbf{CR}   & \textbf{AR}     \\ \hline
\textbf{No Persona}                  & 72.22             & -       & -      & 44.67             & -        & -       & 48.37            & -       & -      \\
\textbf{Scientist}                   & 72.45             & 43.06   & 1.16   & \textbf{47.11}    & 28.00    & 0.78    & \textbf{50.22}   & 18.52   & 0.76   \\
\textbf{Biologist}                   & \textbf{75.31}    & 76.54   & 1.85   & 46.00             & 68.33    & 4.00    & 46.73            & 64.92   & 2.94   \\
\textbf{Chemist}                     & 74.92             & 59.41   & 1.46   & 45.00             & 38.00    & 1.78    & 49.46            & 35.95   & 1.09   \\
\textbf{Physicist}                   & 75.08             & 40.43   & 0.93   & 46.00             & 15.78    & 0.78    & 49.02            & 11.22   & 0.76   \\
\textbf{Artist}                      & 72.76             & 78.16   & 2.31   & 45.22             & 64.00    & 1.67    & 44.12            & 66.56   & 2.61   \\ \hline
\textbf{Scientific (Brief)}          & 73.76             & 13.74   & 0.54   & 46.11             & 5.67     & 0.00    & 47.60            & 4.14    & 0.33   \\
\textbf{Realistic (Brief)}           & 74.69             & 23.46   & 0.54   & 46.67             & 14.66    & 0.22    & 47.82            & 8.61    & 0.33   \\
\textbf{Investigative (Brief)}       & 74.61             & 54.86   & 1.08   & 44.44             & 45.33    & 2.89    & 45.31            & 37.91   & 2.29   \\
\textbf{Artistic (Brief)}            & 72.38             & 66.20   & 1.77   & 45.89             & 54.67    & 2.67    & 46.84            & 57.19   & 2.72   \\
\textbf{Social (Brief)}              & 71.30             & 12.96   & 0.39   & 45.00             & 11.45    & 0.45    & 47.93            & 9.37    & 0.44   \\
\textbf{Enterprising (Brief)}        & 72.91             & 49.15   & 1.31   & 45.00             & 36.22    & 1.11    & 45.97            & 31.70   & 1.63   \\
\textbf{Conventional (Brief)}        & 74.38             & 17.98   & 0.23   & 44.44             & 12.11    & 0.67    & 47.72            & 7.41    & 0.22   \\ \hline
\textbf{Realistic (Descriptive)}     & 73.54             & 32.33   & 0.23   & 44.78             & 21.89    & 0.56    & 44.23            & 14.38   & 0.33   \\
\textbf{Investigative (Descriptive)} & 73.00             & 43.05   & 0.77   & 43.11             & 35.89    & 1.11    & 45.53            & 24.73   & 1.09   \\
\textbf{Artistic (Descriptive)}      & 70.76             & 67.90   & 0.85   & 44.22             & 54.22    & 1.89    & 42.59            & 57.08   & 3.05   \\
\textbf{Social (Descriptive)}        & 70.68             & 71.37   & 1.39   & 43.33             & 61.78    & 2.33    & 44.99            & 61.98   & 3.16   \\
\textbf{Enterprising (Descriptive)}  & 72.76             & 68.90   & 1.00   & 42.78             & 58.78    & 1.45    & 44.55            & 55.66   & 2.18   \\
\textbf{Conventional (Descriptive)}  & 70.99             & 37.96   & 0.85   & 43.78             & 27.22    & 1.33    & 45.21            & 21.79   & 0.65   \\ \hline
\end{tabular}
    \caption{Results of Llama-3-8b.}
    \label{tab:8b}
\end{table*}

\begin{table*}[t!]
\centering
\begin{tabular}{l|ccc|ccc|ccc}
\hline
                            & \multicolumn{3}{c|}{\textbf{College Biology}} & \multicolumn{3}{c|}{\textbf{College Chemistry}} & \multicolumn{3}{c}{\textbf{College Physics}} \\ \hline
\textbf{Persona}                    & \textbf{Acc}   & \textbf{CR}  & \textbf{AR}  & \textbf{Acc}    & \textbf{CR}     & \textbf{AR}     & \textbf{Acc}   & \textbf{CR}   & \textbf{AR}      \\ \hline
\textbf{No Persona}                  & \textbf{71.53} & -     & -    & 42.33          & -     & -    & 44.12          & -     & -    \\
\textbf{Scientist}                   & 67.52          & 2.01  & 0.00 & 40.22          & 0.56  & 0.00 & 43.90          & 1.85  & 0.33 \\
\textbf{Biologist}                   & 70.83          & 5.94  & 0.46 & 40.89          & 15.33 & 1.00 & 40.74          & 33.44 & 4.36 \\
\textbf{Chemist}                     & 70.76          & 5.01  & 0.31 & \textbf{46.78} & 4.11  & 0.00 & \textbf{44.45} & 5.56  & 0.87 \\
\textbf{Physicist}                   & 67.82          & 9.80  & 0.54 & 42.89          & 3.00  & 0.11 & 42.49          & 2.29  & 0.33 \\
\textbf{Artist}                      & 68.44          & 17.28 & 0.93 & 37.11          & 20.44 & 2.89 & 40.53          & 15.69 & 1.96 \\ \hline
\textbf{Scientific (Brief)}          & 69.60          & 0.62  & 0.00 & 41.00          & 0.74  & 0.00 & 42.05          & 0.44  & 0.00 \\
\textbf{Realistic (Brief)}           & 66.12          & 0.15  & 0.00 & 42.22          & 0.00  & 0.00 & 42.27          & 0.00  & 0.00 \\
\textbf{Investigative (Brief)}       & 68.67          & 0.54  & 0.00 & 41.56          & 0.33  & 0.00 & 42.70          & 0.54  & 0.11 \\
\textbf{Artistic (Brief)}            & 69.06          & 15.89 & 1.70 & 43.00          & 17.11 & 2.00 & 43.14          & 18.96 & 2.72 \\
\textbf{Social (Brief)}              & 68.83          & 0.31  & 0.00 & 43.22          & 0.44  & 0.00 & 42.81          & 0.11  & 0.00 \\
\textbf{Enterprising (Brief)}        & 68.44          & 0.93  & 0.00 & 41.44          & 0.89  & 0.00 & 41.51          & 0.98  & 0.22 \\
\textbf{Conventional (Brief)}        & 67.51          & 0.00  & 0.00 & 42.78          & 0.00  & 0.00 & 42.59          & 0.00  & 0.00 \\ \hline
\textbf{Realistic (Descriptive)}     & 46.76          & 1.08  & 0.08 & 34.78          & 1.78  & 0.11 & 34.09          & 0.98  & 0.11 \\
\textbf{Investigative (Descriptive)} & 52.16          & 0.77  & 0.00 & 37.22          & 0.89  & 0.00 & 36.49          & 0.44  & 0.11 \\
\textbf{Artistic (Descriptive)}      & 53.70          & 1.62  & 0.15 & 36.78          & 2.44  & 0.45 & 33.22          & 2.72  & 0.00 \\
\textbf{Social (Descriptive)}        & 51.78          & 0.54  & 0.00 & 36.56          & 0.22  & 0.11 & 33.00          & 0.22  & 0.00 \\
\textbf{Enterprising (Descriptive)}  & 50.46          & 0.54  & 0.08 & 36.67          & 0.11  & 0.11 & 35.73          & 0.11  & 0.00 \\
\textbf{Conventional (Descriptive)}  & 51.47          & 0.62  & 0.08 & 37.33          & 0.22  & 0.00 & 35.62          & 0.65  & 0.00      \\ \hline
\end{tabular}
    \caption{Results of Llama-3-70b.}
    \label{tab:70b}
\end{table*}

\end{document}